\documentclass[letterpaper, 10 pt, conference]{ieeeconf}
\IEEEoverridecommandlockouts
\usepackage[colorlinks,bookmarksopen,bookmarksnumbered,citecolor=red,urlcolor=red]{hyperref}

\usepackage{booktabs,multirow}
\usepackage{multicol}
\usepackage{moreverb,url}
\usepackage{amsfonts}
\usepackage{amsmath}
\usepackage{multirow}
\usepackage{todonotes}
\usepackage{booktabs}
\usepackage{bm}
\usepackage{algpseudocode}
\usepackage{makecell}

\usepackage{algorithm}
\usepackage[english]{babel}
\usepackage{adjustbox}
\DeclareGraphicsExtensions{.pdf,.jpg,.jpeg,.png}
\usepackage{subfig}
\usepackage[font=small,labelfont=bf]{caption}

\usepackage{empheq}

\newcommand\BibTeX{{\rmfamily B\kern-.05em \textsc{i\kern-.025em b}\kern-.08em
T\kern-.1667em\lower.7ex\hbox{E}\kern-.125emX}}
\newcommand{\trsp}{{\scriptscriptstyle\top}}

\usepackage[acronym,shortcuts]{glossaries}

\usepackage{color, colortbl}
\usepackage{bm}
\usepackage{amssymb}
\usepackage{physics}
\usepackage{tabularx,ragged2e,booktabs,caption}
\usepackage{glossaries}

\newacronym{dofs}{DoFs}{Degrees-of-Freedoms}

\graphicspath{{./figures/}} 

\title{\LARGE \bf D-LGP: Dynamic Logic-Geometric Program\\for Reactive Task and Motion Planning
}

\author{Teng Xue$^{1, 2}$, Amirreza Razmjoo$^{1, 2}$, and Sylvain Calinon$^{1, 2}$ 
	\thanks{$^{1}$Idiap Research Institute, Martigny, Switzerland (e-mail: firstname.lastname@idiap.ch)} 
\thanks{$^{2}$École Polytechnique Fédérale de Lausanne (EPFL), Switzerland}%
}

\begin{document}

\maketitle
\thispagestyle{empty}
\pagestyle{empty}

\begin{abstract}
Many real-world sequential manipulation tasks involve a combination of discrete symbolic search and continuous motion planning, collectively known as combined task and motion planning (TAMP). However, prevailing methods often struggle with the computational burden and intricate combinatorial challenges, limiting their applications for online replanning in the real world. To address this, we propose Dynamic Logic-Geometric Program (D-LGP), a novel approach integrating Dynamic Tree Search and global optimization for efficient hybrid planning. Through empirical evaluation on three benchmarks, we demonstrate the efficacy of our approach, showcasing superior performance in comparison to state-of-the-art techniques. We validate our approach through simulation and demonstrate its reactive capability to cope with online uncertainty and external disturbances in the real world.
\\ Project webpage: \href{https://sites.google.com/view/dyn-lgp}{https://sites.google.com/view/dyn-lgp}.
\end{abstract}

\section{Introduction}
\label{sec:introduction}

Robots are increasingly playing a vital role in real-world scenarios, particularly with long-horizon manipulation tasks. These tasks can usually be decoupled with various subtasks, each associated with continuous trajectories. This refers to a problem called combined Task and Motion Planning (TAMP) \cite{garrett2021integrated}, where high-level discrete task planning and low-level continuous motion planning are tightly interconnected. The effectiveness of the motion planner heavily relies on the task skeleton, and, simultaneously, the task planner should be aware of how its outputs are evaluated by the motion planner to avoid spending excessive time on bad solutions. Due to this inherent coupling, the entire system, encompassing all objects and robots, must be considered holistically. 


\begin{figure}[t]
	\centering
	\includegraphics[width=0.7\columnwidth]{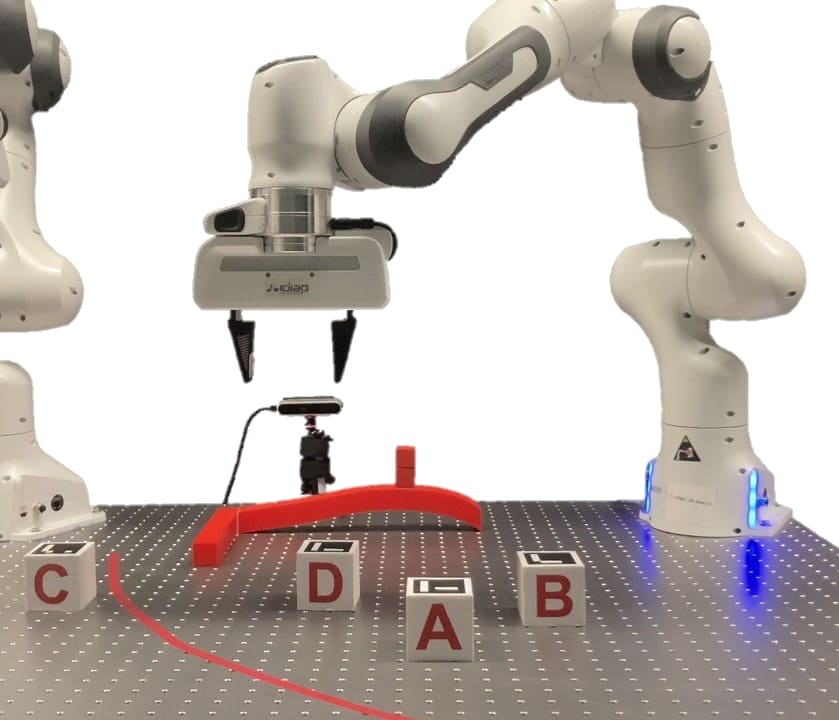}
	\caption{System setup where the robot constructs a tower sequentially. The objective is to stack the blocks at a specified target point in a predefined order. If a block is out of reach (indicated by the red curve), the robot must determine how to use a tool to pull the block into the reachable region before continuing the pick-and-place operation.}
	\label{fig:real_setup}
	\vspace{-0.5cm}
\end{figure}

Many advances have been made to address this integration challenge in TAMP, typically involving a search-based strategy to identify task skeletons and sampling-based or optimization-based motion planning methods to evaluate them. The interplay between the task and motion domains results in significant high-dimensional combinatorial complexity, particularly as the number of objects and the length of trajectory horizon increase. Unlike continuous planning, where dimensionality can be easily reduced thanks to the interpolation nature, such as using basis functions \cite{razmjoo2021optimal}, there is no direct numerical relationship between adjacent symbolic variables in TAMP. This means that even a small change in the high-level task skeleton can lead to a significant difference in the continuous domain. Consequently, this combinatorial complexity of TAMP leads to unavoidable high-dimensional solution space, posing significant challenges for the high-level task planner to find a feasible skeleton from a huge amount of choices. For example, popular discrete planning methods like Monte Carlo Tree Search (MCTS) suffer a lot from the sparse solution space,  despite efforts to introduce multiple bounds \cite{toussaint2017multi} or approximated cost-to-go functions \cite{braun2022rhh}. Low-level motion planning, typically sampling-based or optimization-based, is used to evaluate the quality of the selected action skeleton. Nevertheless, these evaluations often lack guarantees of reliability. Sampling-based techniques are probabilistically complete given infinite time, which is impractical for TAMP to be solved in real time. They yield feasible but non-optimal solutions, while optimality is important for real-world manipulation, such as time and energy efficiency. Optimization-based approaches are geared towards optimality, they primarily rely on nonlinear programming (NLP) \cite{toussaint2015logic}. The resulting local optima can sometimes be satisfactory, but they may lack the confidence to provide certain feedback about feasibility to high-level task planners. In summary, the intricate high-dimensional combinatorial complexities lead to inefficient task sampling, exacerbated by the poor performance of low-level motion planning. These challenges collectively classify TAMP as an NP-hard problem, rendering the quest for a feasible solution time-consuming. This hinders the online replanning for robot manipulation, which is important for robots to interact with the real world reactively.

In this work, we address these challenges by introducing a method that uses backpropagation for high-level action skeleton reasoning and global optimization for low-level motion planning. The task planner employs the optimal principles of Dynamic Programming \cite{bellman1966dynamic}, where the action skeleton is derived through backpropagation from the target configuration. This backward mechanism also eliminates limitations on horizon lengths. Consequently, we refer to it as Dynamic Tree Search (DTS). In terms of low-level motion planning, we consider non-convex motion planning as a combination of many convex subspaces, allowing us to formulate the problem as mixed-integer convex optimization and solve it globally using modern solvers.

Our contributions in this paper include:

1) We propose Dynamic Tree Search (DTS) for high-level task planning, utilizing backpropagation to quickly find the feasible action skeleton from combinatorial exploded solution space.

2) We propose an efficient LGP framework called Dynamic Logic-Geometric Program (D-LGP), combining DTS and global optimization to solve TAMP problems efficiently and optimally.

3) Given the fast computation of D-LGP, we formulate it in closed-loop fashion that is reactive to uncertainty and external disturbances in real-world scenarios.

We evaluated our method on three benchmarks, each comprising multiple subtasks with varying numbers of objects. Our results, both numerical and experimental, demonstrate the efficiency and optimality of the proposed approach.

\section{Related Work}
\label{sec:related_works}

Task and motion planning is initiated with single-flow approaches like aSyMov \cite{cambon2009hybrid} and SMAP \cite{plaku2010sampling}, that start from task planning to motion planning. These methods require a complete description of constraints but lack interplay between the task layer and the motion layer. To address this limitation, more TAMP methods have emerged, allowing low-level motion planners to validate the feasibility of task skeletons. Examples include PDDLStream \cite{garrett2020pddlstream} and FFRob \cite{garrett2018ffrob}. These methods incorporate geometric constraints as logical predicates, demanding extensive human efforts in logic design. The incorporation of human expertise introduces apprehensions regarding the algorithm completeness.


Among the methods mentioned above, \cite{garrett2020pddlstream, garrett2018ffrob} primarily fall into the category of sampling-based TAMP, utilizing existing sampling planners like RRTs \cite{lavalle1998rapidly} to assess the geometric feasibility of selected action skeletons. Sampling-based methods theoretically achieve probabilistic completeness given unlimited time \cite{dantam2018incremental}. However, TAMP typically requires fast motion planning to inform task planners. Moreover, sampling-based methods prioritize feasibility over optimality, while the latter is vital for efficient robot manipulation in terms of time and energy. 

This leads to the emergence of optimization-based TAMP, with Logic-Geometric Programming (LGP) \cite{toussaint2015logic} being a prominent framework. LGP employs Monte-Carlo Tree Search (MCTS) in the STRIPS domain to identify action skeleton candidates and subsequently solves nonlinear trajectory optimization problems with path and mode switching constraints. However, LGPs often exhibit slow runtimes due to the number of calls to the NLP solver. In cases where the search space is sparse, most selected action skeletons are usually infeasible, resulting in a protracted and unsatisfactory exploration process. Several variants have been proposed to mitigate this. For instance, Toussaint et al. introduced multi-bound tree search (MBTS) \cite{toussaint2017multi}, which solves a relaxed NLP problem that serves as a lower bound for the original problem and a necessary condition for feasibility. Building upon this concept, more variants have been put forward, each employing different bounds and heuristics to prune infeasible branches or approximate cost-to-go functions of MCTS. Examples include stability bounds \cite{hartmann2020robust}, prefix conflicts \cite{ortiz2022conflict}, and action-specific heuristics \cite{braun2022rhh}. However, these bounds and heuristics are often domain-specific and depend on human knowledge. Even when well-designed, these search strategies may still struggle with the sparsity of the solution space, leading to the exploration of irrelevant action skeletons and unavoidable time consumption. Unless the cost-to-go functions can be perfectly approximated, which is usually not the case, these challenges persist. Overall, MCTS-based strategies are highly effective for most tasks when given sufficient time. However, they are usually too general to capture the causal structure of certain TAMP problems, particularly those with precisely known target configurations. Such problems are prevalent in the field of robot manipulation, including tabletop rearrangement and assembly tasks.

Our approach falls under the LGP category and aims to achieve efficient and globally optimal TAMP for robot manipulation tasks in which the target configuration is known. We employ backward search to fully uncover the causal structure of long-horizon tasks, enhancing the efficiency of task planning. The feasibility checking adopts the multi-bound concept from \cite{toussaint2017multi} for fast computation. We formulate motion planning problems as mixed-integer convex optimization, efficiently solvable with modern solvers, such as Gurobi \cite{optimization2014inc}, to obtain global optimal solutions or determine infeasibility. This hierarchical structure results in a fast LGP framework, making it possible to be formulated in closed-loop fashion. The whole framework therefore exhibits reactive behaviors at both task and motion levels.

\section{Method}
\label{sec:method}

In this work, we primarily focus on tabletop rearrangement tasks, but our approach can be extended to various tasks where the target configuration is known. We will first present the new target-centric LGP formulation, followed by the mixed-integer expression of motion planning. Finally, we will introduce the complete D-LGP algorithm, encompassing Dynamic Tree Search and full path optimization.

\subsection{Target-centric LGP}
\label{sec:TC-LGP_formu}
The full path of LGP is composed of $K$ phases of a fixed duration $T$. Instead of checking feasibility and computing the cost of full path after finding the full skeleton candidate, we propose to use backward propagation from the target configuration. This requires a new target-centric LGP formulation:
\begin{equation}
	\begin{small}
	\label{eq:TC-LGP}
	\begin{aligned}
		\min_{a_{k}, \bm{x}(t), \bm{u}(t)} \quad & \int_{(k-1)T}^{kT} c(\bm{x}(t), \bm{u}(t)) \, dt + c_T (\bm{x}(kT), g_k)\\
		\textrm{s.t.} \quad & \bm{x}(0) = \bm{x}_0, s_{k} \in \mathcal{G} (g_k), g_K = g_K \\
		&h_{\text{goal}}(\bm{x}(kT), g_k) = 0
		\\
  		&h_{\text{path}}(\bm{x}(t), \bm{u}(t) | s_k(t)) = 0 \\ & g_{\text{path}}(\bm{x}(t), \bm{u}(t) | s_k(t)) \leq 0 \\
		&h_{\text{switch}}(\bm{x}(kT) | a_k, s_{k-1}) = 0     \\ & g_{\text{switch}}(\bm{x}(kT) \, | \, a_k, s_{k-1}) \leq 0 \\
		& a_k \in \mathcal{A} (s_{k-1}), s_k \in succ(s_{k-1}, a_k) \\
		& g_{k-1} \in \mathcal{T}^\dagger (g_k),	 
	\end{aligned}
	\end{small}
\end{equation}
where $\bm{x}(t)$ and $\bm{u}(t)$ are the continuous trajectory and control variables, with $\bm{x}(kT)$ as the final configuration of phase $k$. $h_{\text{path}}$ and $g_{\text{path}}$ represent the path constraints, and $h_{\text{switch}}$, $g_{\text{switch}}$ denote the transition conditions between two phases with the action operator $a_k$. $g_k$ is the target to be achieved at phase $k$. $s_{k} \in \mathcal{G} (g_k)$ and $h_{\text{goal}}(\cdot, \cdot)$ describes the goal specification logically and geometrically. $c(.)$ and $c_T(.)$ are the path cost and terminal cost to penalize control effort (e.g., energy, force) and the final geometric performance. The logical state transition from $s_{k-1}$ to $s_k$ is determined by $succ(., .)$ as a function of $s_{k-1}$ and action $a_k \in \mathcal{A} (s_{k-1})$. $\mathcal{T}^\dagger (g_k)$ identifies the essential subgoals to attain $g_k$, which is described in detail in Alg. \ref{alg:TaskGraph}. The formulation \eqref{eq:TC-LGP} adopts the concepts and variables from the standard LGP formulation \cite{toussaint2015logic}, while allowing dynamic programming for backpropagation from the target configuration $g_K$. This mechanism can be viewed as spatial reasoning over the causal structure of the task.

\subsection{Mixed-integer Motion Planning}
\label{sec:MIP} 

The goal of tabletop rearrangement is to manipulate blocks to achieve a desired configuration. This task involves a set of objects denoted as $\mathcal{O}$, each with poses represented as $P_n^t = (x_n^t, y_n^t, \theta_n^t) \in SE(2)$ at time $t$ for object $n$. In essence, the problem entails determining the optimal placement $\bm{u}_t = [x_j^t \; y_j^t]^\trsp$ for a selected object $j$. This task is notably challenging due to the non-convex nature of the solution set. Rather than attempting to solve the problem within this non-convex set, we adopt a strategy of decomposition into multiple convex sets. Subsequently, the constraints for each object $n \in 1, \dots, N$ can be succinctly expressed as combinations of four halfspaces (we assume the objects are square blocks), denoted by $i \in \{1, 2, 3, 4\}$:

\begin{equation}
\begin{small}
	\label{eq:halfsapce}
	\begin{aligned}
		C_i: \quad &[a_i \quad b_i] \; \bm{R} (\theta_n^t)^\trsp \; \bm{u}_t \geq c_i,
	\end{aligned}
\end{small}
\end{equation}
where $\bm{R}(\theta_n^t)$ is a rotation matrix of the block frame and w.r.t. the global frame:
\begin{small}
\begin{equation*} 
	\bm{R}(\theta_n^t) = \begin{bmatrix} \text{cos}(\theta_n^t) & -\text{sin}(\theta_n^t) \\ \text{sin}(\theta_n^t) & \text{cos}(\theta_n^t)  \end{bmatrix},
\end{equation*}
\end{small}

\noindent and $a_1 = 1, b_1 = 0, c_1 = l/2+x_{n, b}^t$, $a_2 = -1, b_2 = 0, c_2 = l/2-x_{n, b}^t$, $a_3 = 0, b_3 = 1, c_3 = l/2+y_{n, b}^t$, $a_4=0, b_4 = -1, c_4 = l/2-y_{n, b}^t$, $[x_{n, b}^t \quad y_{n, b}^t] = [x_{n}^t \quad y_{n}^t] \bm{R} (\theta_n^t)$. $l$ is the size of the square block.

As a consequence, the non-intersection constraints among these four halfspaces can be viewed as a disjunctive constraint $\lor_{i=1}^4 C_i$, which can be further represented by combining constraints along with the introduction of additional binary variables, as discussed in \cite{blackmore2011chance}.
 
The above formulation can be converted as Mixed-Integer Quadratic Program (MIQP) characterized by a quadratic objective and linear constraints. This transformation incorporates binary variables denoted as $z_i$ to determine the specific halfspace within which the placement should reside. In the case of each halfspace, the constraints can be expressed linearly. To articulate the if-else condition related to binary variables, we can employ a conventional big-M formulation, similar to \cite{quintero2023optimal}, to represent \eqref{eq:halfsapce} as 
\begin{equation}
	\begin{small} 
		\label{eq:bigM}
		\begin{aligned}
			 C_n: \quad &[a_i \quad b_i] \; \bm{R}(\theta_n^t)^\trsp \; \bm{u}_t \geq c_i - M (1-z_i),\\
			 &\sum z_i = 1, \quad z_i \in \{0, 1\}, \quad \forall i \in \{1,2,3,4\}.
		\end{aligned}
	\end{small}
\end{equation}

Given the action $a_{k}$, the motion planning problem in \eqref{eq:TC-LGP} can be described as
\begin{equation}
\label{eq:MIQP}
	\begin{small}
	\begin{aligned}
			\min_{\bm{x}(t), \bm{u}(t)} \quad & \int_{(k-1)T}^{kT} c(\bm{x}(t), \bm{u}(t)) \, dt + c_T (\bm{x}(kT), g_k)\\
			\textrm{s.t.} \quad  &\bm{x}(0) = \bm{x}_0, h_{\text{goal}}(\bm{x}(kT), g_k) = 0
			\\
			&\bm{u}(t) \in \cup_{n=1}^N \{C_n\} - \{C_m\},
		\end{aligned}
		\end{small}
\end{equation}
where $m$ represents the block intended to be positioned by action $a_{k}$. Additionally, another similar constraint defined by the geometric goal should be taken into account for the intermediate actions. For instance, as illustrated in Fig. \ref{block:solution}, the action \texttt{Pick[B \#P2]} must ensure that block \texttt{B} does not occupy position \texttt{\#P3} to facilitate subsequent actions.

\subsection{D-LGP Algorithm}
\label{sec:D-LGP}

To tackle \eqref{eq:TC-LGP} iteratively from the target configuration, we introduce D-LGP in Alg. \ref{alg:D-LGP}. Here, we employ a pseudoinverse successor function, denoted as $succ^\dagger(s_k, s_{k+1})$, to deduce the goal-oriented action $a_{k+1}$ for transitioning from $s_k$ to $s_{k+1}$. This function disregards the symbolic and geometric constraints, focusing solely on determining the necessary action to reach the target state $s_{k+1}$ from $s_k$.

\begin{tiny}
\begin{algorithm}[t] 
\caption{Conflict-driven TaskGraph}
\label{alg:TaskGraph}
\begin{algorithmic}[1]
\Require{\\
$\bm{x}(kT):$ current geometric state, $s_k:$ current symbolic state, $g_K:$ target configuration 
}
\Ensure{$ \mathcal{B}_a:$ action skeleton, $ \mathcal{B}_g:$ subgoal skeleton}
\State{\textbf{Initialize:}
\State{$s_K \in \mathcal{G}(g_K)$}
\State{$a_{k+1} = succ^\dagger(s_{k}, s_K)$}
\State{$\mathcal{B}_a=[a_{k+1}], \mathcal{B}_g=[g_K]$}}
\While{$(\bm{x}(kT), a_{k+1})$ infeasible} 
\State{$g^* = sub\_goal(\bm{x}(kT), a_{k+1})$} \Comment{subgoal reasoning}
\State{$s^* \in \mathcal{G}(g^*)$}
\State{$a_{k+1}= succ^\dagger(s_k, s^*)$} 
\State{$\mathcal{B}_a.append(a_{k+1})$}
\State{$\mathcal{B}_g.append(g^*)$}
\EndWhile
\end{algorithmic}
\end{algorithm}
\end{tiny}

\begin{tiny}
\begin{algorithm}[htbp] 
	\caption{D-LGP: Solving LGP using Dynamic Tree Search and Global Optimization}
	\label{alg:D-LGP}
	\begin{algorithmic}[1]
\Require{\\
$\bm{x}_0:$ initial configuration, $s_0:$ initial symbolic state, $g_K:$ target configuration 
}
\Ensure{$ a_{1:K}:$ action skeleton, $\bm{x}(t):$ full path}
		\Statex{\textbf{ $\quad \quad \; \quad$ Level 1: Dynamic Tree Search}}

		{\State{\textbf{Initialize:} \State{$\bm{x}(0) = \bm{x}_0$, $s_0 \in \mathcal{L}$, $k=1$}}
		\State{$\mathcal{B}_a, \mathcal{B}_g \leftarrow \text{TaskGraph}(\bm{x}_0, s_0, g_K)$}
		\While {not solved}
			\State{$a_k = \mathcal{B}_a[-1]$} (feasible action given $\bm{x}_{k-1}$)
			\State{$s_{k} = succ(s_{k-1}, a_k)$}
			\State{$g_k = \mathcal{B}_g[-1]$}
			\State{$\bm{x}(kT), \bm{u}(kT) = \text{MIQP}(a_k, g_k)$} \Comment{Eq. \eqref{eq:MIQP}}

			\State{$\mathcal{B}_{a}, \mathcal{B}_g = \text{TaskGraph}(\bm{x}(kT), s_k, g_K)$  \Comment{see Alg. \ref{alg:TaskGraph}}}
			\State{$k \leftarrow k + 1$}
		\EndWhile}
		\vspace{-4mm}
		\begin{center}
		\Statex{\textbf{Level 2: Full Optimization}}
		\end{center}
		\State{\textbf{Given} $a_{1:K}$:}\\
		\State{$\min \limits_{\bm{x}(t), \bm{u}(t)} \int_0^{KT} c(\bm{x}(t), \bm{u}(t)) \, dt + c_T (\bm{x}(KT), g)$}\\ 
	\end{algorithmic}

\end{algorithm}

\end{tiny}

Given the current configuration $\bm{x}((k-1)T)$, we employ the conflict-driven TaskGraph algorithm (Alg. \ref{alg:TaskGraph}) to assess the feasibility of the proposed action. If the action is found infeasible, we generate a sequence of actions ($\mathcal{B}_a$) and subgoals ($\mathcal{B}_g$) to construct a tree connecting $s_{k-1}$ to $s_{K}$. $sub\_goal(\cdot)$ identifies the crucial subgoals that must be accomplished first to render action $a_{k}$ feasible, given the current configuration $\bm{x}((k-1)T)$. 
Subsequently, we calculate the continuous path from $t=(k-1)T$ to $t=kT$ based on \eqref{eq:MIQP}. Since computing \eqref{eq:MIQP} is expensive, we decide to only focus on the final pose at $t=kT$ for fast feasibility checking and cost computation. Based on this, we can quickly solve it and obtain the placement locations $\bm{u}(kT)$ or returns global infeasibility. 

\begin{figure}[htbp] 
\centering
\subfloat[Initial and target states]{{\includegraphics[width=0.4\columnwidth]{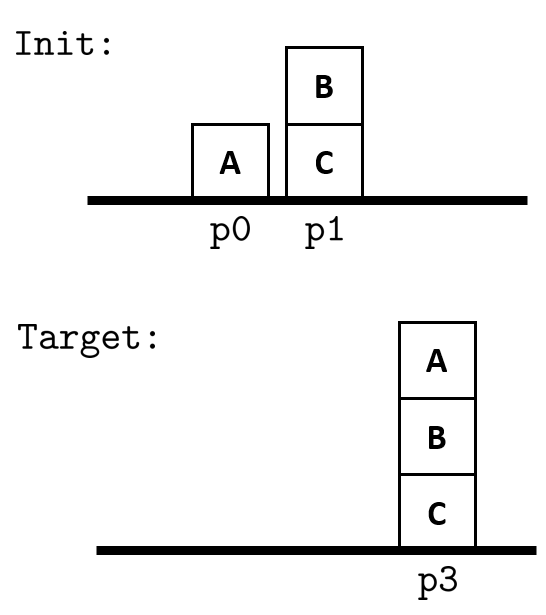}} \label{block:config}}
\subfloat[Solution]{{\includegraphics[width=0.6\columnwidth]{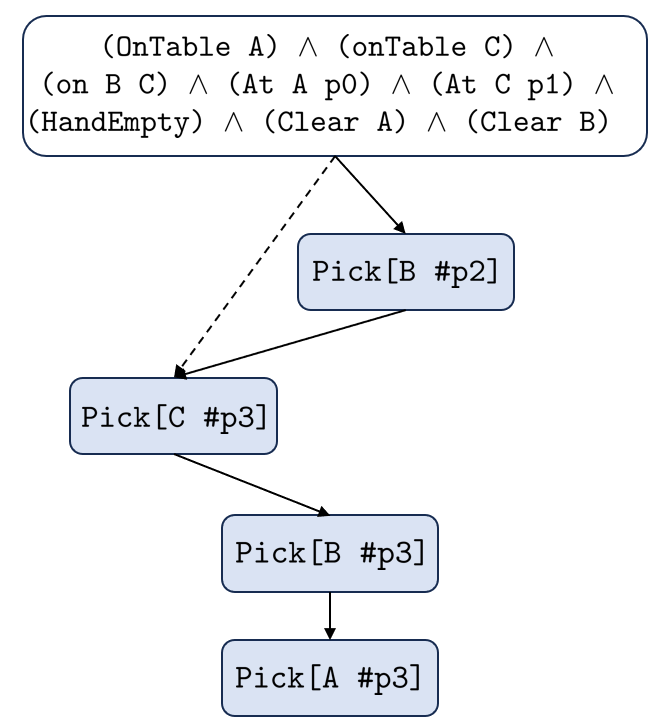}}\label{block:solution} \label{block:solution}} 
	\caption{Illustrative example of Dynamic Tree Search. Given the initial and target configurations, $succ^\dagger(s_0, s_{K})$ will infer the essential action as \texttt{Pick[C \#P3]} (shown as dotted line in Fig. \ref{block:solution}). However, this is infeasible because B is on top of C. $sub\_{goal}(\cdot)$ will identify the essential subgoal as moving B to \texttt{\#p2} (given by MIQP) first. Then, C can be picked and placed at \texttt{\#p3}, followed by picking B and A until reaching the final target.}
	\label{fig:block_world}
	\vspace{-0.5cm}
\end{figure}

Fig. \ref{fig:block_world} presents a simplified example to illustrate the running process of Dynamic Tree Search. Given the initial and target configurations, the essential action to connect them is \texttt{Pick[C \#P3]}, inferred through $succ^\dagger(s_0, s_{K})$ that disregards all the constraints. It is clear that \texttt{Pick[C \#P3]} is infeasible because B is on top of C. $sub\_{goal}(\cdot)$ will then be used to infer the appropriate configuration $g^*$ to make \texttt{Pick C} feasible. This process iterates until finding a feasible action respecting the current geometric configuration. In the provided example, the inferred action is \texttt{Pick B}, which is feasible concerning $\bm{x}(0)$. Then, Eq. \eqref{eq:MIQP} is invoked to generate the optimal placement \texttt{\#p2}, ensuring minimal end-effector movement while satisfying geometric constraints. After executing the action, $s_0$ and $\bm{x}(0)$ update to $s_1$ and $\bm{x}(T)$, respectively. $succ^\dagger(\cdot, \cdot)$ is then used to determine the necessary action to achieve $s_K$ from $s_1$, resulting in \texttt{Pick C}, which is geometrically feasible w.r.t. $\bm{x}(T)$. The process continues similarly for \texttt{Pick [B \#p3]} and \texttt{Pick [A \#p3]}, until reaching the final target configuration $g_K$.

After finding the full action skeleton $\{a_1, \dots, a_K\}$, a full path optimization will be called to find the global optimal trajectory, focusing on the keyframes $t=0, T, \dots, KT$. This idea coincides with the pose bound $\mathcal{P}_{pose}$ and sequence bound $\mathcal{P}_{seq}$ used in multi-bound tree search \cite{toussaint2017multi}, which proves using coarse path discretization can effectively speed up the optimization process. The difference is that we eliminate the full path bound $\mathcal{P}_{full}$ to fully utilize the robot controller to cope with uncertainties and external disturbance in the real world. Our TAMP approach runs in Cartesian space and relies on Operational Space Controller to actuate the robot joints, leading to reactive motion control. This is also highlighted in \cite{migimatsu2020object}, where a Cartesian formulation of LGP was proposed.

\section{Experiments}
\label{sec:experiments}
In this section, we will present the comparison results between our approach and state-of-the-art methods.
\subsection{Benchmarks}
We test our method in 3 different tabletop domains, all requiring long-horizon physical reasoning about the discrete actions and continuous motions. The reward is only given at the termination of the whole process. Across all task domains, we employ a random selection of 15 distinct initializations to assess the performance of our algorithm.

\textbf{a) Obstructed Pick X (OP-X)}:
In a cluttered environment, the robot's task is to grasp an object surrounded by $X-1$ taller objects. If the taller objects are located near the target object, the robot must relocate them to create a clear path for grasping. This necessitates the robot's comprehensive exploration of a long-horizon combinatorial solution space, taking into account the geometric relationships among each object. Furthermore, the placement locations in continuous space must also be considered to minimize end-effector displacement. Choosing the right placement of blocks can have a substantial impact on the later end-effector movement. 

\textbf{b) Tower Construction X (Tower-X)}:
X blocks are placed on the table, either in a stacked or unstacked configuration. The goal is to construct a tower following a specific order. Similar to the Pick-X task, the blocks may have varying heights, which means that a shorter block cannot be picked directly if a taller one is adjacent to it. Additionally, as mentioned in \cite{quintero2023optimal}, both motion-level constraints and task-level constraints should be taken into account. For instance, if block A is initially atop block B, and the target order requires maintaining this arrangement at the target point, then A must be temporarily positioned at an intermediate point to first move block B and then A. Furthermore, in contrast to \cite{quintero2023optimal}, we also consider the overall path optimality. This means that the placement of block B should be considered in the entire trajectory to ensure minimum energy cost.

\textbf{c) Tower Construction X with Tool (Tower-Tool-X)}:
The blocks will be placed on the table within a large area, sometimes beyond the robot's reach. Consequently, the robot must determine whether to use a tool (in the form of a hook) to pull objects into its reachable area before grasping them. This task is an expansion of the Tower-X scenario, incorporating a pulling action in addition to the standard grasping action. It also expands the workspace of a single robot arm to accomplish more complicated tasks. Furthermore, the robot must determine the target of the pulling phase to obtain a full optimal trajectory. We also demonstrate this task in real robot experiments.

\subsection{DTS vs MBTS}
\label{sec:MCTS}

In this section, we compare our Dynamic Tree Search (DTS) with the Multi-Bound Tree Search (MBTS), which is widely used in previous LGP solvers. Our DTS adopts a similar idea of pose bound $\mathcal{P}_{pose}$ for quickly evaluating the feasibility of the current node but utilizes backward search from the target configuration. Therefore, we use $\mathcal{P}_{pose}$ for both MBTS and DTS to ensure a fair comparison. Moreover, considering that the parameter used to balance exploitation and exploration in MCTS/MBTS can significantly impact the results, we compare our DTS approach against three variations of MBTS. Each variant employs a distinct exploration-exploitation strategy, including greedy (MBTS-0), medium (MBTS-1) and lazy (MBTS-2). The comparison results are presented in Table \ref{tab:DTS_vs_MBTS}, with 100 seconds as the computation time threshold. We find that our proposed method visits the fewest nodes in the shortest time and achieves a $100\%$ success rate for all tasks. Moreover, thanks to the backpropagation strategy, it is applicable to tasks with exceedingly long horizons.

\renewcommand{\arraystretch}{1.}
\begin{table*}[htbp]
	\caption{Comparison of the proposed DTS approach with other MBTS approaches}
	\begin{footnotesize}
		\scalebox{0.73}{
		\begin{tabular}{l | c c c | c c c| c c c| c c c|}
			\toprule
			  & \multicolumn{3}{c|}{DTS} & \multicolumn{3}{c|}{MBTS-0}& \multicolumn{3}{c|}{MBTS-1}& \multicolumn{3}{c|}{MBTS-2}\\
		  \cline{2-13}
			  & time (s) &nodes &success  & time (s) &nodes   &success  & time (s) &nodes  &success  & time (s) &nodes  &success  \\
			\midrule
			{OP-6} & \textbf{0.02  $\pm$ 0.01}  &  \textbf{2.8 $\pm$  1.0} &\textbf{100\%} & 0.84  $\pm$ 0.93 & 48.2 $\pm$ 49.0  &100\% & 0.59  $\pm$ 0.40 & 35.5 $\pm$ 21.5  &100\% & 1.00  $\pm$ 1.32 & 59.8 $\pm$ 71.6 &100\% \\
			{OP-9} & \textbf{0.07  $\pm$ 0.04}  &  \textbf{7.1 $\pm$  2.0} &\textbf{100\%} & 10.5  $\pm$ 12.5 & 320.5 $\pm$ 386.4 &100\% & 11.8  $\pm$ 16.1 & 359.2 $\pm$ 480.8  &100\% & 29.0  $\pm$ 35.4 & 949.3 $\pm$ 1140.5 &100\% \\
			{OP-12} & \textbf{0.11  $\pm$ 0.03}  &  \textbf{10.4  $\pm$  2.0}  & \textbf{100\%} & 27.20  $\pm$ 26.98  & 672.6  $\pm$ 674.8 & 33\% & 50.86  $\pm$ 39.07  &  1301.4  $\pm$  1028.5  & 33\% & 32.25  $\pm$ 30.11  & 886.3  $\pm$ 845.2 & 20\%\\
			{OP-15} & \textbf{0.19  $\pm$ 0.05}  &  \textbf{14.1  $\pm$  1.4}  & \textbf{100\%} & -  & - & 0\% & -  & -  & 0\% & - & - & 0\%\\
			{OP-50} & \textbf{5.97  $\pm$ 0.19}  &  \textbf{49.6  $\pm$  0.5}  & \textbf{100\%} & -  & - & 0\% & -  & -  & 0\% & - & - & 0\%\\
			{Tower-4} & \textbf{0.01  $\pm$ 0.01}  &  \textbf{4.7 $\pm$  0.7} &\textbf{100\%} & 0.27  $\pm$ 0.12 & 19.8 $\pm$ 8.1  &100\% & 0.28  $\pm$ 0.12 & 19.8 $\pm$ 6.8  &100\% & 0.91  $\pm$ 0.36 & 71.6 $\pm$ 32.4 &100\% \\
			{Tower-8} & \textbf{0.13  $\pm$ 0.05}  &  \textbf{10.9 $\pm$  1.8} &\textbf{100\%} & 93.04  $\pm$ 10.73 & 4784 $\pm$ 1033 &46.7\% & 97.04  $\pm$ 3.78 & 4917.4 $\pm$ 827.6  &53.3\% &96.62  $\pm$ 3.33 & 4864.6 $\pm$ 879.4 &46.7\% \\
			{Tower-12} &  \textbf{0.15  $\pm$ 0.05}  &   \textbf{17.3  $\pm$  2.64} &  \textbf{100\%}  & -  & - & 0\% & -  & -  & 0\% & - & - & 0\%\\
			{Tower-50} & \textbf{2.28  $\pm$ 0.76}  &  \textbf{75.6  $\pm$  12.1}  & \textbf{100\%} & -  & - & 0\% & -  & -  & 0\% & - & - & 0\%\\
			
			{Tower-Tool-4} & \textbf{0.02  $\pm$ 0.01}  &  \textbf{7.2  $\pm$  0.5} & \textbf{100\%} & 0.80  $\pm$ 0.21  & 33.8  $\pm$ 9.2 & 100\% & 1.00  $\pm$ 0.43  &  44.2  $\pm$  19.2  & 100\% & 2.33  $\pm$ 1.00  & 113.1  $\pm$ 49.5 & 100\%\\
			{Tower-Tool-8} & \textbf{0.14  $\pm$ 0.04}  &  \textbf{15.3  $\pm$  0.6} & \textbf{100\%} & -  & - & 0\% & -  &  -  & 0\% & -  & -5 & 0\%\\
			\bottomrule
		\end{tabular}
	}
	\end{footnotesize}
	\label{tab:DTS_vs_MBTS}
	\vspace{-0.4cm}

\end{table*}

\subsection{MIP vs SLSQP and IPOPT}
\label{sec:SLSQP}

Due to the coupling structure between high-level task planning and low-level motion planning, the quality of low-level results can significantly influence the search direction of the task planner. In this paper, we aim to demonstrate the substantial benefits of global optimization in solving LGPs compared to widely used NLP solvers such as IPOPT and SLSQP. We present the comparison results in Table \ref{tab:motion_solver}.

To obtain these results, we employ DTS as the high-level task planner and subsequently utilize MIQP, IPOPT, and SLSQP for low-level motion planning on the benchmarks with an increasing number of blocks. The implementation is based on Gurobi \cite{optimization2014inc}, CasADi \cite{andersson2019casadi}, and Scipy \cite{virtanen2020scipy}, respectively. The objective of motion planning here is to determine the placement locations of the picked object. This task is particularly challenging due to the non-convex nature of the set of feasible solutions. As more obstacles are introduced, the problem becomes increasingly complex.

Successful planning is defined as obtaining a feasible solution within 10 seconds. We can observe from Table \ref{tab:motion_solver} that for setups with a small number of objects, all three solvers can easily achieve a high success rate within a short computation time. However, the success rates of the two NLP solvers decrease dramatically when more blocks are involved, indicating that NLP solvers struggle with complex non-convex problems. In contrast, MIQP can still provide optimal results. This highlights the fact that, given a feasible action skeleton, NLP solvers may be trapped in poor local optima or even fail to obtain a feasible solution. Consequently, the high-level task planner receives incorrect guidance, leading to a failure to identify the genuinely feasible sequence. Global optimization is therefore important to facilitate the communication between task planner and motion planner.


\renewcommand{\arraystretch}{1.}
\begin{table}[htbp]
	\caption{Comparison of MIQP, IPOPT and SLSQP}
	\begin{footnotesize}
		\scalebox{0.75}{
			\begin{tabular}{l | c c | c c| c c|}
				\toprule
				& \multicolumn{2}{c|}{MIQP} & \multicolumn{2}{c|}{IPOPT}& \multicolumn{2}{c|}{SLSQP}\\
				\cline{2-7}
				& time (s) &success  & time (s) &success  & time (s) &success \\
				\midrule
				{OP-3} & 0.01  $\pm$ 0.01 &100\% & 0.03  $\pm$ 0.07 &100\% & \textbf{0.01  $\pm$ 0.00}  &\textbf{100\%}  \\
				{OP-6} & 0.02  $\pm$ 0.02 &100\% & 0.04  $\pm$ 0.07 &100\% & \textbf{0.02  $\pm$ 0.01} &\textbf{100\%} \\
				{OP-9} & \textbf{0.12  $\pm$ 0.05} &\textbf{100\%} & 0.07  $\pm$ 0.02 &46.7\% & 0.32  $\pm$ 0.23 &40.0\%  \\
				{OP-12} & \textbf{0.18  $\pm$ 0.04} &\textbf{100\%} & 0.09  $\pm$ 0.03 &20.0\% & -   &0\%  \\
				{OP-15} & \textbf{0.22  $\pm$ 0.04} &\textbf{100\%} & 0.18  $\pm$ 0.0 &6.7\% & -   &0\%  \\
				{OP-50} & \textbf{5.95  $\pm$ 0.32} &\textbf{100\%} & - &0\% & -   &0\%  \\
				{Tower-4} & \textbf{0.04  $\pm$ 0.02} &\textbf{100\%} & 0.18  $\pm$ 0.16 &100\% & 0.07  $\pm$ 0.01 &100\%  \\
				{Tower-8} & \textbf{0.07  $\pm$ 0.03} &\textbf{100\%} & 0.10  $\pm$ 0.08 &86.7\% & 0.20  $\pm$ 0.20 &53.3\%  \\
				{Tower-12} & \textbf{0.15  $\pm$ 0.08} &\textbf{100\%} & 0.12  $\pm$ 0.03 &46.7\% & 0.42  $\pm$ 0.34 &60.0\%  \\
				{Tower-50} & \textbf{2.33  $\pm$ 0.85} &\textbf{100\%} &- &0\% & - &0\%  \\
				{Tower-Tool-4} & 0.04  $\pm$ 0.02 &100\% & \textbf{0.02  $\pm$ 0.01} &\textbf{100\%} & 0.06  $\pm$ 0.04 &100\%  \\
				{Tower-Tool-8} & \textbf{0.10  $\pm$ 0.03} &\textbf{100\%} & 0.07  $\pm$ 0.02 &93.3\% & 0.27  $\pm$ 0.44 &66.7\%  \\
				\bottomrule
			\end{tabular}
		}
	\end{footnotesize}
	\label{tab:motion_solver}
		\vspace{-0.2cm}
	
\end{table}

\subsection{D-LGP Performance}
\label{sec:LGP}

We conducted a detailed analysis of our proposed algorithm on various benchmarks. The results are summarized in Table \ref{tab:D-LGP}. Our findings indicate that using $\mathcal{P}_{pose}$ for DTS significantly accelerates the hybrid programming, resulting in a feasible action skeleton and suboptimal motion variables. Given that TAMP involves long-horizon planning to achieve the lowest cost along the entire path, we subsequently perform full optimization based on the generated task sequence. When comparing the end-effector displacement between DTS and full optimization levels, we observed that full optimization can indeed yield an optimal full trajectory, whereas DTS assumes prior movements to be perfect and aims to rapidly find the optimal solution for the current step. However, it is worth noting that full optimization requires more computational time. This is primarily due to the significant increase of binary variables, which poses challenges for modern MIQP solvers like Gurobi. With a greater number of objects, full optimization may encounter difficulties. This can be solved in the future by employing Iterative Regional Inflation by Semidefinite programming (IRIS) \cite{deits2015computing} to simplify MIQP formulation. Additionally, we provide the makespan of the plans, which closely aligns with the number of objects involved. This demonstrates that our generated plans prioritize executing actions that are essential rather than including feasible but extraneous actions. 

\renewcommand{\arraystretch}{1.}
\begin{table}[htbp]
	\caption{Statistical analysis of D-LGP performance}
	\begin{footnotesize}
		\scalebox{0.67}{
			\begin{tabular}{l |c c | c c| c c|}
				\toprule
				& \multicolumn{2}{c|}{DTS} & \multicolumn{2}{c|}{Full Optimization}& \multicolumn{2}{c|}{Total}\\
				\cline{2-7}
				& time(s) &EE Disp.(m)  & time(s) &EE Disp.(m)  & total(s) &makespan \\
				\midrule
				{OP-6} &0.03  $\pm$ 0.02  &0.49  $\pm$ 0.12 & 0.22  $\pm$ 0.11 &0.42 $\pm$ 0.07 &0.25  $\pm$ 0.12 &3.2  $\pm$ 1.4 \\
				{OP-9} &0.10  $\pm$ 0.05 &0.73  $\pm$ 0.24 & 1.39  $\pm$ 1.75 &0.65 $\pm$ 0.15 &1.49  $\pm$ 1.79 &5.47  $\pm$ 1.93 \\
				{OP-12} &0.13  $\pm$ 0.01 &1.26  $\pm$ 0.09 & 5.93  $\pm$ 2.93 &0.93 $\pm$ 0.10 &6.05 $\pm$ 2.92 &7.0  $\pm$ 0.82 \\
				{Tower-4} &0.02  $\pm$ 0.01 &1.74  $\pm$ 0.49 & 0.08  $\pm$ 0.02 &1.70 $\pm$ 0.45 &0.10  $\pm$ 0.03 &4.33  $\pm$ 0.47 \\
				{Tower-8} &0.09  $\pm$ 0.04 &2.41 $\pm$ 0.54 & 0.52  $\pm$ 0.22 &2.22 $\pm$ 0.49 &0.62  $\pm$ 0.24 &10.4  $\pm$ 1.82 \\
				{Tower-12} &0.16  $\pm$ 0.07 &3.19  $\pm$ 0.63 & 0.83  $\pm$ 0.16 &3.03 $\pm$ 0.46 &0.99  $\pm$ 0.22 &13.4  $\pm$ 1.36 \\
				{Tower-Tool-4} &0.03  $\pm$ 0.02 &1.10  $\pm$ 0.17 & 0.02  $\pm$ 0.02 &0.86 $\pm$ 0.14 &0.05  $\pm$ 0.03 &6.73  $\pm$ 0.57 \\
				{Tower-Tool-8} &0.08  $\pm$ 0.02 &3.30  $\pm$ 0.57 & 0.44  $\pm$ 1.02 &2.72 $\pm$ 0.54 &0.52  $\pm$ 1.01 &15.0  $\pm$ 0.52 \\
				\bottomrule
			\end{tabular}
		}
	\end{footnotesize}
	\label{tab:D-LGP}
	\vspace{-0.5cm}
	
\end{table}
\begin{figure*}[htbp] 
	\centering
	\subfloat[Initialization]{{\includegraphics[width=0.29\columnwidth]{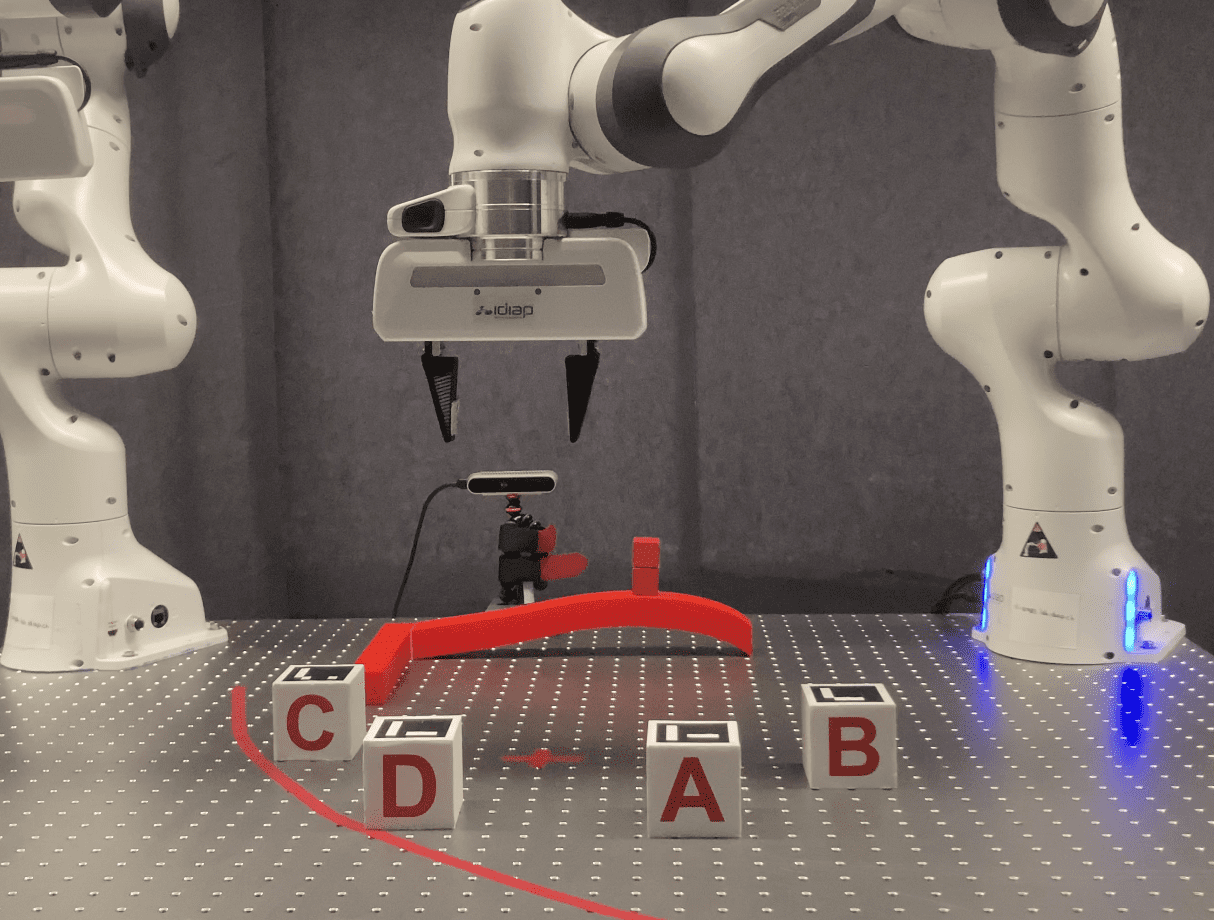}}}
	\subfloat[Place D]{{\includegraphics[width=0.29\columnwidth]{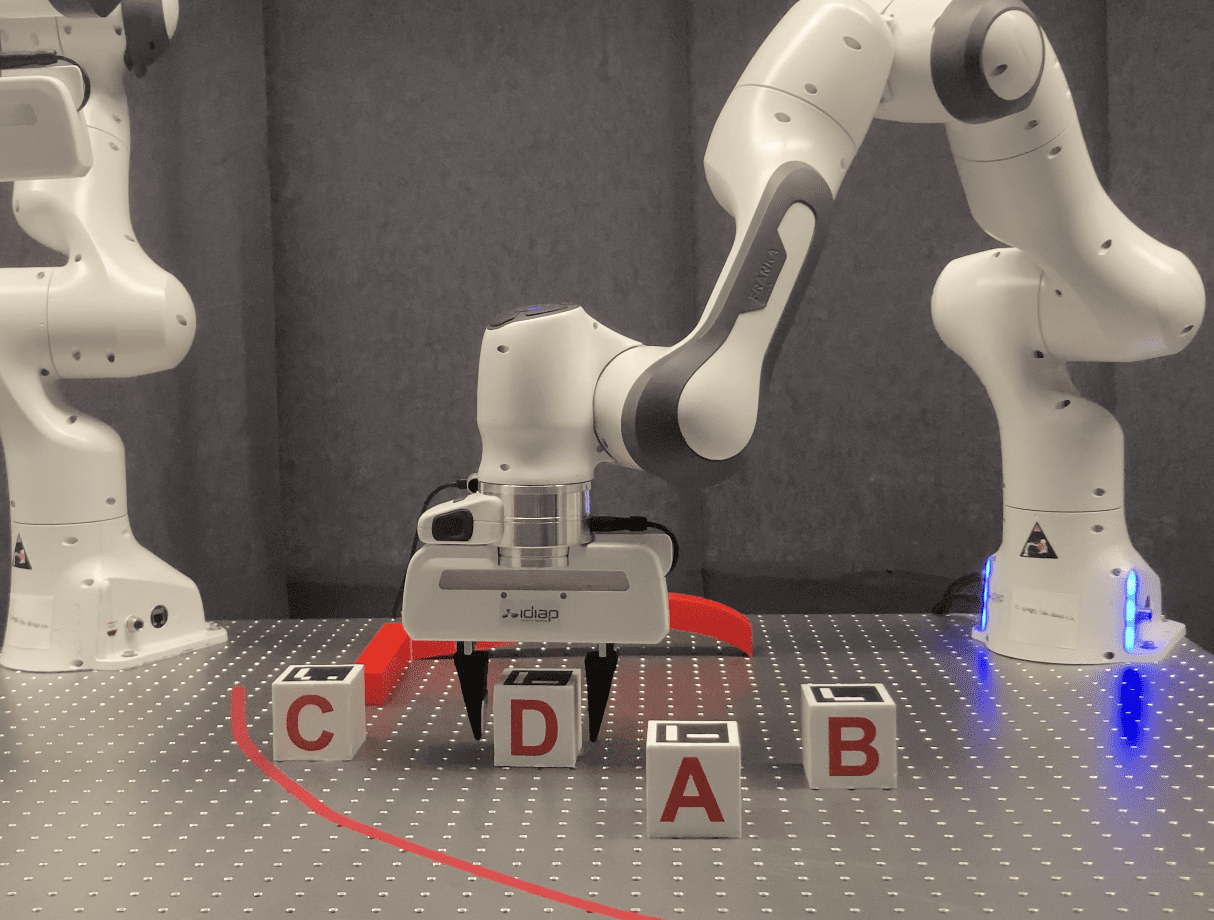}}} 
	\subfloat[Tool usage]{{\includegraphics[width=0.29\columnwidth]{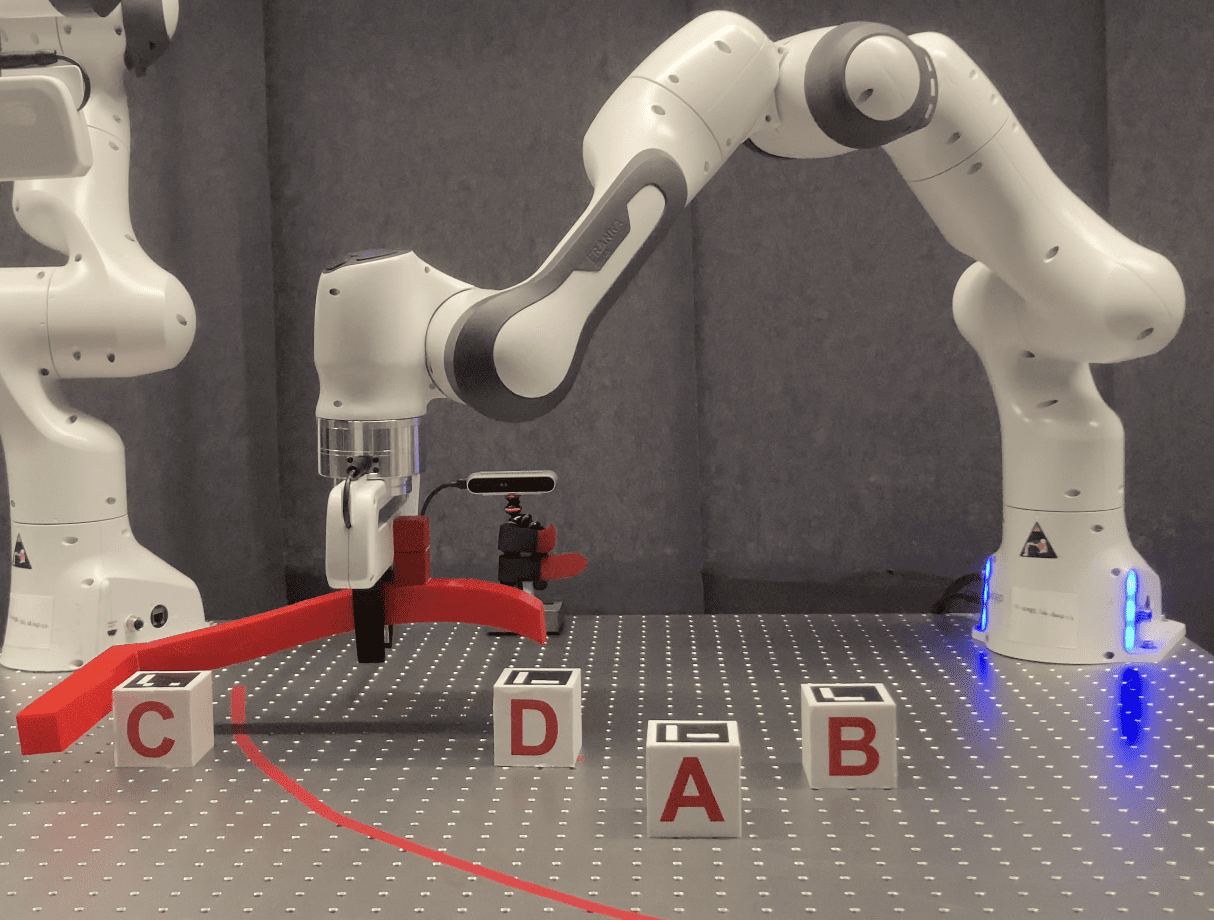}}} 
	\subfloat[Pulling]{{\includegraphics[width=0.29\columnwidth]{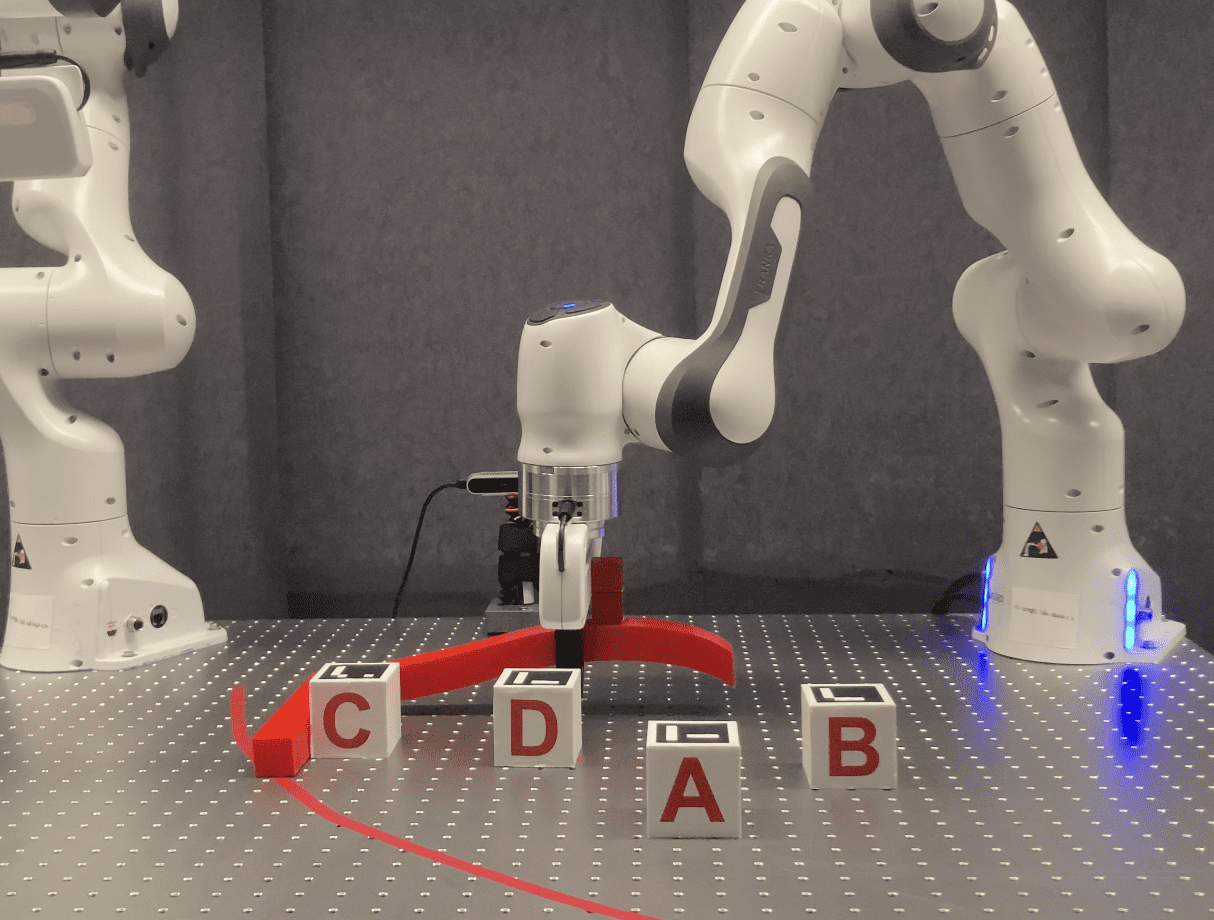}}} 
	\subfloat[Place C]{{\includegraphics[width=0.29\columnwidth]{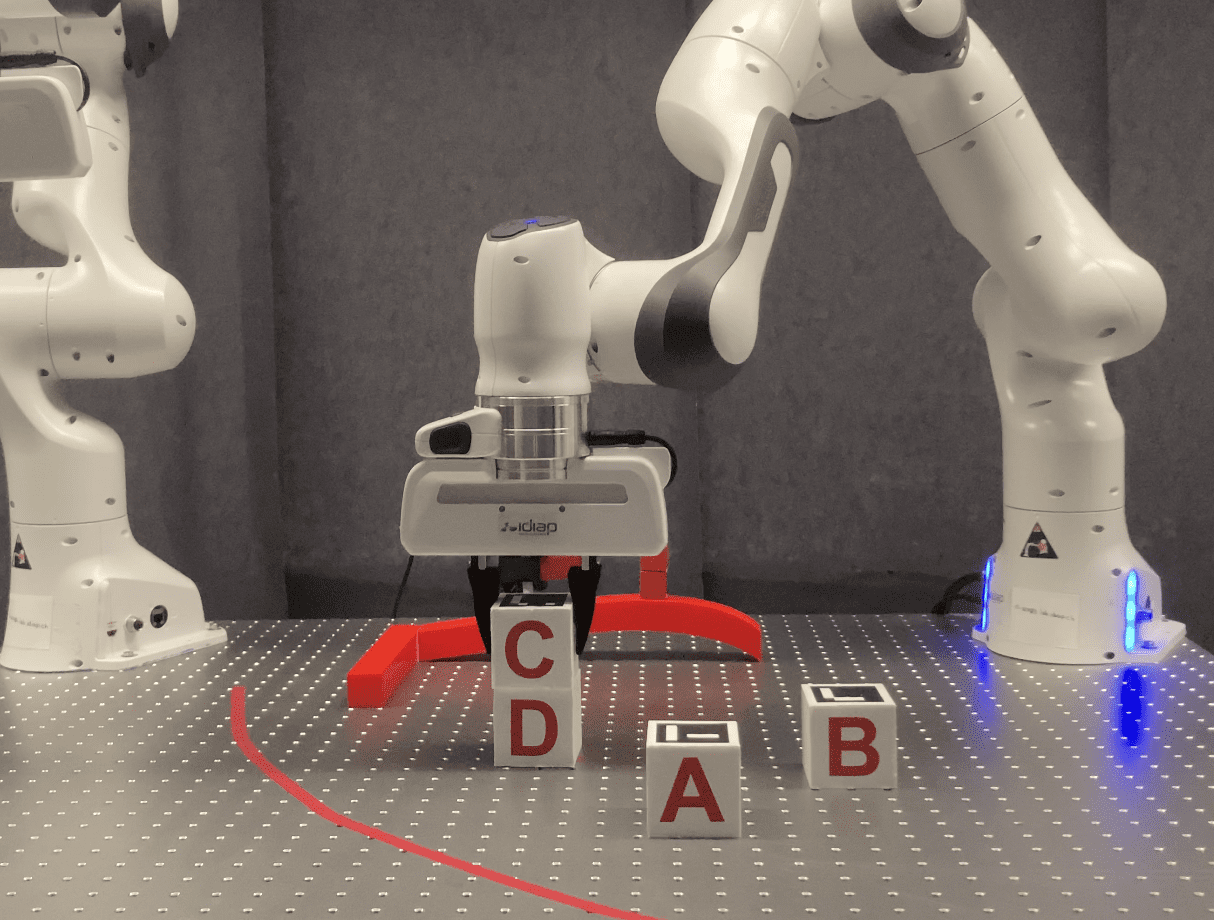}}} 
	\subfloat[Place B]{{\includegraphics[width=0.29\columnwidth]{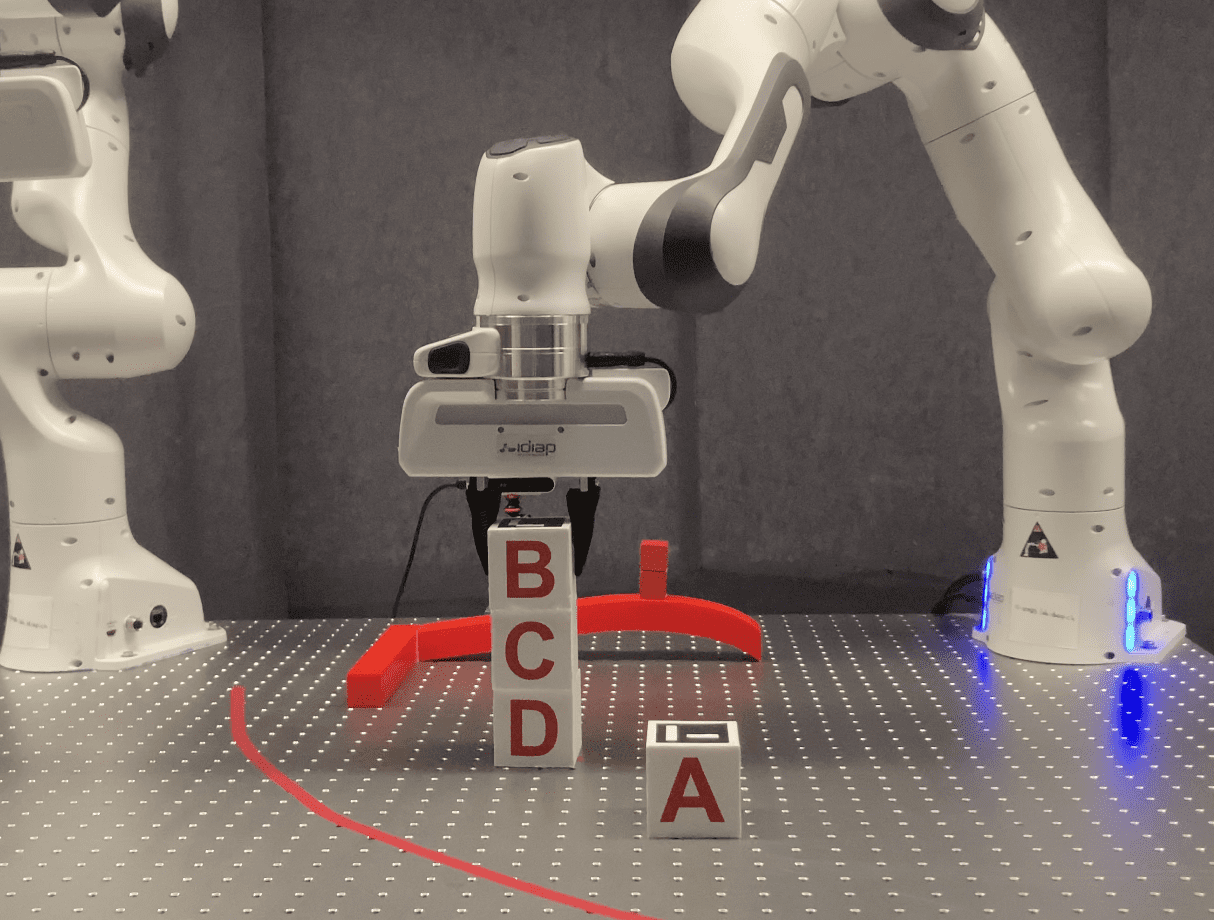}}}
	\subfloat[Place A]{{\includegraphics[width=0.29\columnwidth]{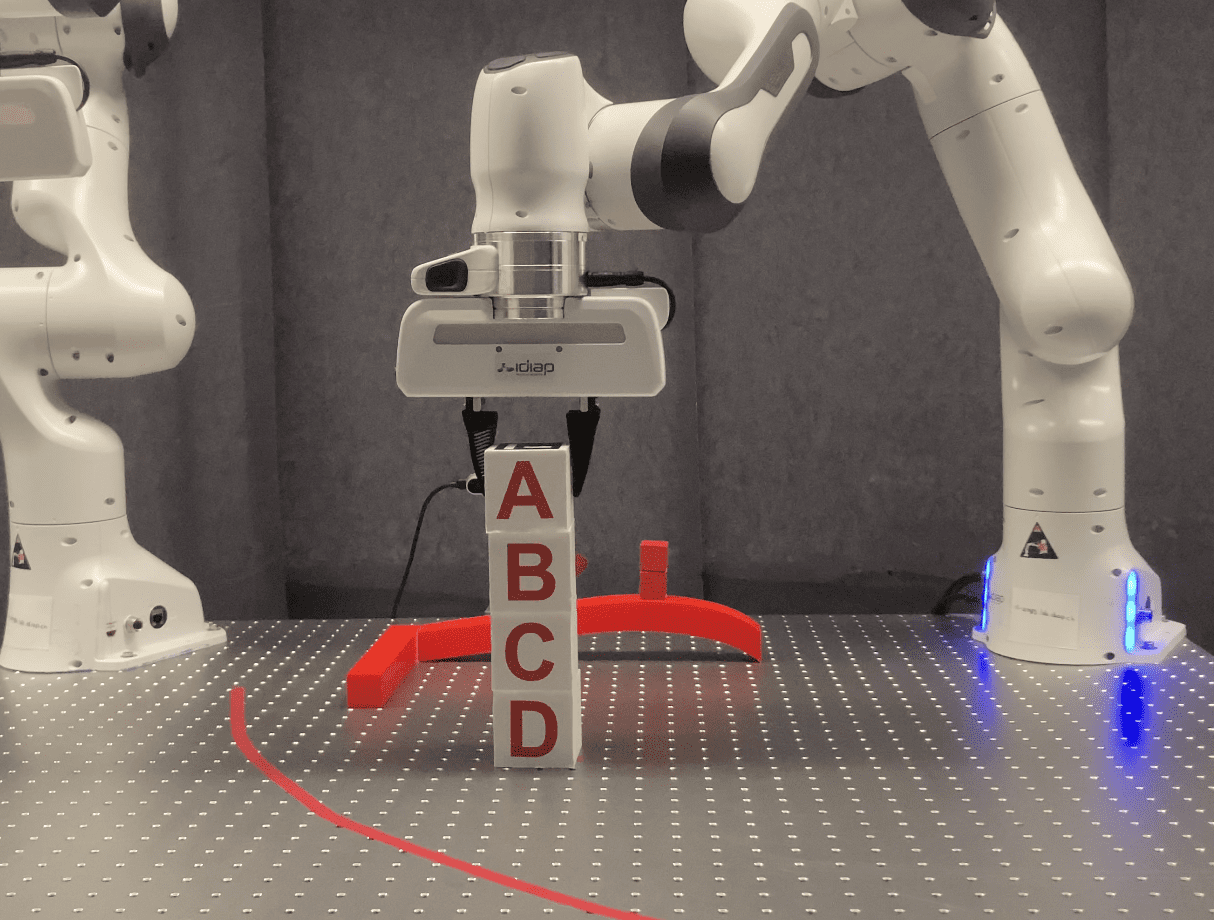}}} 
	\caption{Tower construction task with tool usage. The configuration is initialized as (a) and the objective is to stack blocks as (g). The first step is to pick and place D at the target point (b). Then, an external disturbance occurs, moving block C out of reachability. To resolve this, the robot uses a tool (c) to pull block C into the reachability region (d). Once done, the robot can pick C as usual and place it on top of D (c). Next, B (f) and A (g) are placed in sequence.}
	\label{fig:keyframe}
	\vspace{-0.5cm}
\end{figure*}

\subsection{Real robot experiments}
\label{sec:real_robot}

We conducted tests on a 7-axis Franka Emika robot (Fig. 1), utilizing a RealSense D435 camera for object localization. The objective was to stack blocks in a predetermined order at a designated target point. The system started from a random configuration, from which the robot autonomously determined a long-horizon task sequence and the corresponding motion trajectory. Additionally, if any block was out of the robot's reach, it autonomously employed a hook to bring it closer before picking it up.

As mentioned in Section \ref{sec:D-LGP}, the output of D-LGP is defined in the object-centric frame, enabling adaptive low-level adjustments. For example, if the location of the target block changes, the robot's end-effector can continue tracking it until picking it up. Moreover, since D-LGP runs significantly faster compared to standard LGP solvers, we have formulated it in closed-loop fashion. The D-LGP solver runs at approximately 10Hz, generating long-horizon task and motion sequences. We apply the first output to the robot while continuing to run D-LGP with the new state obtained from the 30Hz camera. This framework enables high-level task replanning in response to inaccurate execution (e.g., if a block falls) and external disturbances (e.g., random shifts in stacked block orders). We demonstrate these reactive behaviors in our accompanying video.

\section{Conclusion and Future Work}
\label{sec:conclusion}

In this paper, we introduce a novel LGP framework called D-LGP, which leverages Dynamic Tree Search (DTS) and global optimization for reactive TAMP. We demonstrate that DTS enables target-oriented search and eliminates constraints on horizon length, allowing for the handling of tasks with extended horizons. We also show that our integrated global optimization formulation is capable of quickly obtaining global optima if the problem is feasible, evaluating the output of DTS effectively. We tested this approach on various benchmarks, highlighting its strong performance compared to state-of-the-art methods.

In this work, we used separate integer variables to represent each subspace corresponding to every face of the object. The number of integer variables grows significantly with the number of objects, making it impractical to solve. This issue could be mitigated by utilizing a single integer variable to represent each convex region in the obstacle-free space.

Currently, our framework is tailored for tasks with explicit target configurations and does not support implicit target descriptions, like ``build the tower as high as possible''. However, we believe that our method can be adapted for such cases by combining it with other approaches to determine the target configuration first.

Moreover, we believe that the proposed method could be integrated into Model-based Reinforcement Learning (RL) approaches, particularly in scenarios with a sparse reward setup to facilitate efficient long-horizon exploration.


\newpage

\IEEEtriggeratref{30}
%
\bibliographystyle{IEEEtran}
\bibliography{main}

\begin{thebibliography}{10}
\providecommand{\url}[1]{#1}
\csname url@rmstyle\endcsname
\providecommand{\newblock}{\relax}
\providecommand{\bibinfo}[2]{#2}
\providecommand\BIBentrySTDinterwordspacing{\spaceskip=0pt\relax}
\providecommand\BIBentryALTinterwordstretchfactor{4}
\providecommand\BIBentryALTinterwordspacing{\spaceskip=\fontdimen2\font plus
\BIBentryALTinterwordstretchfactor\fontdimen3\font minus
  \fontdimen4\font\relax}
\providecommand\BIBforeignlanguage[2]{{%
\expandafter\ifx\csname l@#1\endcsname\relax
\typeout{** WARNING: IEEEtran.bst: No hyphenation pattern has been}%
\typeout{** loaded for the language `#1'. Using the pattern for}%
\typeout{** the default language instead.}%
\else
\language=\csname l@#1\endcsname
\fi
#2}}

\bibitem{garrett2021integrated}
C.~R. Garrett, R.~Chitnis, R.~Holladay, B.~Kim, T.~Silver, L.~P. Kaelbling, and
  T.~Lozano-P{\'e}rez, ``Integrated task and motion planning,'' \emph{Annual
  review of control, robotics, and autonomous systems}, vol.~4, pp. 265--293,
  2021.

\bibitem{razmjoo2021optimal}
A.~Razmjoo, T.~S. Lembono, and S.~Calinon, ``Optimal control combining
  emulation and imitation to acquire physical assistance skills,'' in
  \emph{2021 20th International Conference on Advanced Robotics (ICAR)}.\hskip
  1em plus 0.5em minus 0.4em\relax IEEE, 2021, pp. 338--343.

\bibitem{toussaint2017multi}
M.~Toussaint and M.~Lopes, ``Multi-bound tree search for logic-geometric
  programming in cooperative manipulation domains,'' in \emph{Proc. {IEEE} Intl
  Conf. on Robotics and Automation ({ICRA})}, 2017, pp. 4044--4051.

\bibitem{braun2022rhh}
C.~V. Braun, J.~Ortiz-Haro, M.~Toussaint, and O.~S. Oguz, ``Rhh-lgp: Receding
  horizon and heuristics-based logic-geometric programming for task and motion
  planning,'' in \emph{Proc. {IEEE/RSJ} Intl Conf. on Intelligent Robots and
  Systems ({IROS})}, 2022, pp. 13\,761--13\,768.

\bibitem{toussaint2015logic}
M.~Toussaint, ``Logic-geometric programming: an optimization-based approach to
  combined task and motion planning,'' in \emph{Proceedings of the 24th
  International Conference on Artificial Intelligence}, 2015, pp. 1930--1936.

\bibitem{bellman1966dynamic}
R.~Bellman, ``Dynamic programming,'' \emph{Science}, vol. 153, no. 3731, pp.
  34--37, 1966.

\bibitem{cambon2009hybrid}
S.~Cambon, R.~Alami, and F.~Gravot, ``A hybrid approach to intricate motion,
  manipulation and task planning,'' \emph{Intl Journal of Robotics Research},
  vol.~28, no.~1, pp. 104--126, 2009.

\bibitem{plaku2010sampling}
E.~Plaku and G.~D. Hager, ``Sampling-based motion and symbolic action planning
  with geometric and differential constraints,'' in \emph{Proc. {IEEE} Intl
  Conf. on Robotics and Automation ({ICRA})}, 2010, pp. 5002--5008.

\bibitem{garrett2020pddlstream}
C.~R. Garrett, T.~Lozano-P{\'e}rez, and L.~P. Kaelbling, ``Pddlstream:
  Integrating symbolic planners and blackbox samplers via optimistic adaptive
  planning,'' in \emph{Proceedings of the International Conference on Automated
  Planning and Scheduling}, vol.~30, 2020, pp. 440--448.

\bibitem{garrett2018ffrob}
C.~R. Garrett, T.~Lozano-Perez, and L.~P. Kaelbling, ``Ffrob: Leveraging
  symbolic planning for efficient task and motion planning,'' \emph{Intl
  Journal of Robotics Research}, vol.~37, no.~1, pp. 104--136, 2018.

\bibitem{lavalle1998rapidly}
S.~LaValle, ``Rapidly-exploring random trees: A new tool for path planning,''
  \emph{Research Report 9811}, 1998.

\bibitem{dantam2018incremental}
N.~T. Dantam, Z.~K. Kingston, S.~Chaudhuri, and L.~E. Kavraki, ``An incremental
  constraint-based framework for task and motion planning,'' \emph{Intl Journal
  of Robotics Research}, vol.~37, no.~10, pp. 1134--1151, 2018.

\bibitem{hartmann2020robust}
V.~N. Hartmann, O.~S. Oguz, D.~Driess, M.~Toussaint, and A.~Menges, ``Robust
  task and motion planning for long-horizon architectural construction
  planning,'' in \emph{Proc. {IEEE/RSJ} Intl Conf. on Intelligent Robots and
  Systems ({IROS})}, 2020, pp. 6886--6893.

\bibitem{ortiz2022conflict}
J.~Ortiz-Haro, E.~Karpas, M.~Toussaint, and M.~Katz, ``Conflict-directed
  diverse planning for logic-geometric programming,'' in \emph{Proceedings of
  the International Conference on Automated Planning and Scheduling}, vol.~32,
  2022, pp. 279--287.

\bibitem{optimization2014inc}
\BIBentryALTinterwordspacing
{Gurobi Optimization, Inc.}, ``Gurobi optimizer reference manual,'' 2014.
  [Online]. Available: \url{http://www.gurobi.com/}
\BIBentrySTDinterwordspacing

\bibitem{blackmore2011chance}
L.~Blackmore, M.~Ono, and B.~C. Williams, ``Chance-constrained optimal path
  planning with obstacles,'' \emph{IEEE Transactions on Robotics}, vol.~27,
  no.~6, pp. 1080--1094, 2011.

\bibitem{quintero2023optimal}
C.~Quintero-Pena, Z.~Kingston, T.~Pan, R.~Shome, A.~Kyrillidis, and L.~E.
  Kavraki, ``Optimal grasps and placements for task and motion planning in
  clutter,'' in \emph{Proc. {IEEE} Intl Conf. on Robotics and Automation
  ({ICRA})}, 2023, pp. 3707--3713.

\bibitem{migimatsu2020object}
T.~Migimatsu and J.~Bohg, ``Object-centric task and motion planning in dynamic
  environments,'' \emph{{IEEE} Robotics and Automation Letters ({RA-L})},
  vol.~5, no.~2, pp. 844--851, 2020.

\bibitem{andersson2019casadi}
J.~A. Andersson, J.~Gillis, G.~Horn, J.~B. Rawlings, and M.~Diehl, ``Casadi: a
  software framework for nonlinear optimization and optimal control,''
  \emph{Mathematical Programming Computation}, vol.~11, no.~1, pp. 1--36, 2019.

\bibitem{virtanen2020scipy}
P.~Virtanen, R.~Gommers, T.~E. Oliphant, M.~Haberland, T.~Reddy, D.~Cournapeau,
  E.~Burovski, P.~Peterson, W.~Weckesser, J.~Bright, \emph{et~al.}, ``Scipy
  1.0: fundamental algorithms for scientific computing in python,''
  \emph{Nature methods}, vol.~17, no.~3, pp. 261--272, 2020.

\bibitem{deits2015computing}
R.~Deits and R.~Tedrake, ``Computing large convex regions of obstacle-free
  space through semidefinite programming,'' in \emph{Algorithmic Foundations of
  Robotics XI: Selected Contributions of the Eleventh International Workshop on
  the Algorithmic Foundations of Robotics}.\hskip 1em plus 0.5em minus
  0.4em\relax Springer, 2015, pp. 109--124.

\end{thebibliography}

\end{document}